\documentclass{amsart}
\usepackage{amsaddr}
\usepackage{graphicx}
\usepackage{hyperref}

\begin{document}

\title{Searching for a practical evidence of 
\\the No Free Lunch theorems}

\author{Mihai Oltean}

\address{Department of Computer Science,\\
Faculty of Mathematics and Computer Science,\\
Babes-Bolyai University, Kogalniceanu 1,\\
Cluj-Napoca, 3400, Romania.\\
\url{https://mihaioltean.github.io}
}
\email{mihai.oltean@gmail.com}

\maketitle

\begin{abstract} According to the \textit{No Free Lunch} (NFL) theorems all black-box algorithms perform equally well when compared over the entire set of optimization problems. An important problem related to NFL is finding a test problem for which a given algorithm is better than another given algorithm. Of high interest is finding a function for which Random Search is better than another standard evolutionary algorithm. In this paper we propose an evolutionary approach for solving this problem: we will evolve 
test functions for which a given algorithm $A$ is better than another given 
algorithm $B$. Two ways for representing the evolved functions are employed: as GP trees and as binary strings. Several numerical experiments involving NFL-style Evolutionary Algorithms for function optimization are performed. The results show the effectiveness of the proposed approach. Several test functions for which Random Search performs better than all other considered algorithms have been evolved.

\end{abstract}

\section{Introduction}

Since the advent of the No Free Lunch (NFL) theorems in 1995 \cite{wolpert1,wolpert2}, the 
trend of Evolutionary Computation (EC) \cite{goldberg1} have not changed at all, 
although these breakthrough theories should have produced dramatic changes. 
Most researchers chose to ignore NFL theorems: they developed new 
algorithms that work better than the old ones on some particular test 
problems. The researchers have eventually added: \\

"The algorithm $A$ performs 
better than another algorithm \textit{on the considered test functions}". \\

That is somehow useless since the proposed algorithms cannot be the best on all the considered test functions. 

Moreover, most of the functions employed for testing algorithms are 
artificially constructed. 

Consider for instance, the field of evolutionary single-criteria 
optimization where most of the algorithms were tested and compared on some 
artificially constructed test functions (most of them being known as De'Jong 
test problems) \cite{goldberg1,yao1}. These test problems were used for comparison 
purposes before the birth of the NFL theorems and they are used even today 
(9 years later after the birth of the NFL theorems). Evolutionary 
multi-criteria optimization was treated in a similar manner: most of the 
recent algorithms in this field were tested on several artificially 
constructed test functions proposed by K. Deb in \cite{deb1}.

Roughly speaking, the NFL theorems state that all black-box optimization 
algorithms perform equally well over the entire set of optimization 
problems. Thus, if an algorithm $A$ is better than another algorithm $B$ on some 
classes of functions, the algorithm $B$ is better than $A$ on the rest of the 
functions.

As a consequence of the NFL theories, even a computer program (implementing 
an Evolutionary Algorithm (EA)) containing programming errors can perform 
better than some other highly tuned algorithms for some test functions.

Random Search (RS) being a black box search / optimization algorithm 
should perform better than all of the other algorithms for some classes of 
test functions. Even if this statement is true, there is no result reported -- in 
the literature -- of a test function for which RS performs better 
than all the other algorithms (taking into account the NFL restriction 
concerning the number of distinct solutions visited during the search). 
However, in \cite{droste1} is presented a function which is hard for all Evolutionary Algorithms.

Instead, a lot effort is spent for proving that the No Free Lunch theorems are not true. Most researchers \cite{droste1,droste2,streeter1,whitley1,whitley2} have tried to find some classes of problems for which  NFL does not hold. For instance, in \cite{droste2} is shown that NFL might not hold for small problems (that have a small search space). 

Three questions (on how we match problems to algorithms) are of high interest:

\begin{itemize}

\item{For a given class of problems, what is (are) the algorithm(s) that performs 
(perform) better than all other algorithms?}

\item{For a given algorithm what is (are) the class(es) of problems for which the 
algorithm performs best?}

\item{Given two algorithms $A$ and $B$, what is (are) the class (es) of problems for 
which $A$ performs better than $B$?}

\end{itemize}

Answering these questions is not an easy task. All these problems are still 
open questions and they probably lie in the class of the NP-Complete 
problems. If this assumption is true it means that we do not know if we are 
able to construct a polynomial algorithm that takes a function as input and 
outputs the best optimization algorithm for that function (and vice versa). 
Fortunately, we can try to develop a heuristic algorithm able to handle this 
problem.

In this paper we develop a framework for constructing test functions that 
match a given algorithm. More specific, given two algorithms $A$ and $B$, the 
question is: 

What the functions for which $A$ performs better than $B$ (and vice-versa) are? 

For obtaining such functions we will use an evolutionary 
approach: the functions matched to a given algorithm are evolved by using an standard evolutionary algorithms \footnote{Source code used for evolving test function is available at www.nfl.cs.ubbcluj.ro}.

Of high interest is finding a test function for which Random Search performs better than all  considered standard evolutionary algorithms. Using the proposed approach we were able to evolve such test problems.

The paper is organized as follows: The NFL algorithm is minutely described 
in section \ref{nfl}. Test functions represented as GP trees are evolved in section \ref{real_representation}. The fitness assignment process is described in section \ref{real_fit}. The algorithms used for comparison are described in section \ref{real_acceptance}. Several numerical experiments are carried out in section \ref{real_experiments}. Test functions represented as binary strings are evolved in section \ref{binary_representation}. The fitness assignment process is described in section \ref{binary_fit}. The algorithms used for comparison are described in section \ref{binary_acceptance}. Several numerical experiments are carried out in section \ref{binary_experiments}.

\section{A NFL-style Algorithm}\label{nfl}

We define a black-box optimization algorithm as indicated by 
Wolpert and McReady in \cite{wolpert1,wolpert2}.

The evolutionary model (the NFL-style algorithm) employed in this paper 
uses of a population consisting of a single individual. This considerably simplifies the description and the implementation of a NFL-style algorithm. 

No archive for storing the best solutions found so far (see for instance \textit{Pareto Archived Evolution Strategy} \cite{knowles1}) is maintained. However, we implicitly maintain an archive containing all the distinct solutions explored until the current state. We 
do so because only the number of distinct solutions is counted in the NFL 
theories. This kind of archive is also employed by Tabu Search \cite{glover1,glover2}.

The variables and the parameters used by a NFL-style algorithm are given in Table \ref{tab1}.

\begin{table}[htbp]
\caption{The variables used by the NFL algorithm.}
\label{tab1}
\begin{center}
\begin{tabular}
{p{60pt}p{200pt}}
\hline
\textbf{Variable}& 
\textbf{Meaning} \\
\hline
\textit{Archive}& 
the archive storing all distinct solutions visited by algorithm \\
\textit{curr{\_}sol}& 
the current solution (point in the search space) \\
\textit{new{\_}sol}& 
a new solution (obtained either by mutation or by initialization) \\
\textit{MAXSTEPS}& 
the number of generations (the number of distinct points in the search space visited by the algorithm). \\
$t$& 
the number of distinct solutions explored so far \\
\hline
\end{tabular}
\end{center}
\end{table}

The algorithm starts with a randomly generated solution (the current solution) over the search space. This solution is added to the archive. The following steps are executed until MAXSTEPS different solutions are explored: Generate a solution in the neighborhood of the current solution. This new solution is usually obtained by mutating the current solution. We have to ensure that the newly generated solution is different from all previously explored solutions (The algorithm that generates a solution different from all other solutions explored so far is given further in this section). We add the generated solution to the archive and it becomes the current solution which will be further explored.

The NFL-style algorithm is the following:\\

\newpage

\begin{center}
\textbf{NFL-style Algorithm}
\end{center}

\textsf{\textit{S}}$_{1}$\textsf{. Archive = $\emptyset $;}

\textsf{\textit{S}}$_{2}$\textsf{. Randomly initializes the current solution 
(curr{\_}sol)}

\hspace{1cm}\textsf{// add the current solution to the archive}

\textsf{\textit{S}}$_{3}$\textsf{. Archive = Archive + {\{}curr{\_}sol{\}}; 
}

\textsf{\textit{S}}$_{4}$\textsf{. t = 1;}

\textsf{\textit{S}}$_{5}$\textsf{. }\textsf{\textbf{while}}\textsf{ t $<$ 
MAXSTEPS }\textsf{\textbf{do}}\textsf{ }

\textsf{\textit{S}}$_{6}.$\hspace{1cm}\textsf{ Select a new solution (new{\_}sol) in }

\textsf{the neighborhood of the curr{\_}sol}

\textsf{\textit{S}}$_{7}.$\hspace{1cm}\textsf{ Archive = Archive + {\{}new{\_}sol{\}};}

\textsf{\textit{S}}$_{8}.$\hspace{1cm}\textsf{ curr{\_}sol = new{\_}sol;}

\textsf{\textit{S}}$_{9}.$\hspace{1cm}\textsf{ t = t + 1;}

\textsf{\textit{S}}$_{10}$\textsf{. }\textsf{\textbf{endwhile}}\\

An important issue concerning the NFL algorithm described above is related 
to the step $S_{6}$ which selects a new solution that does not belong to the 
\textit{Archive}. This is usually done by mutating the current solution and keeping the 
offspring if the latter does not already belong to the \textit{Archive} (The actual 
acceptance mechanism is minutely described in sections \ref{real_acceptance} and \ref{binary_acceptance}). If the offspring 
belongs to the \textit{Archive} for a fixed number of mutations (steps) it means that the 
neighborhood of the current solutions could be exhausted (completely 
explored). In this case, a new random solution is generated and the search 
process moves to another region of the search space. It is sometimes possible that the generated solution to already belong to the \textit{Archive}. In this case, another random solution is generated over the search space. We assume that the search space is large enough and after a finite number of re-initializations the generated solution will not belong to the \textit{Archive}.

The algorithm for selecting a new solution which does not belong to the 
\textit{Archive} (the step $S_{6})$ is given below:\\

\textsf{\textit{SS}}$_{1}.$\textsf{ nr{\_}mut = 0; // the number of mutations is set to 0}

\textsf{\textit{SS}}$_{2}.$\textsf{ }\textsf{\textbf{Repeat}}

\textsf{\textit{SS}}$_{3}.$\hspace{1cm}textsf{ new{\_}sol = }\textsf{\textit{Mutate}}\textsf{ (curr{\_}sol);}

\textsf{\textit{SS}}$_{4}.$\hspace{1cm}\textsf{ nr{\_}mut = nr{\_}mut + 1;}

\textsf{\textit{SS}}$_{5}$\textsf{. }\textsf{\textbf{until}}\textsf{ 
(nr{\_}mut = MAX{\_}MUTATIONS) and (new{\_}sol $ \notin 
$}\textsf{\textit{Archive}}\textsf{) }\textsf{\textbf{and}}\textsf{ 
}\textsf{\textit{Accepted}}\textsf{(new{\_}sol);}

\textsf{\textit{SS}}$_{6}$\textsf{. }\textsf{\textbf{while}}\textsf{ 
new{\_}sol $ \notin $Archive }\textsf{\textbf{do}}

\textsf{\textit{SS}}$_{7.}$\hspace{1cm}\textsf{\textit{Initialize}}\textsf{(new{\_}sol); 
//we jump in another randomly chosen point of the search space}

\textsf{\textit{SS}}$_{8}$\textsf{. }\textsf{\textbf{endwhile}}\\

\section{Real-valued functions}\label{real_representation}

Test functions represented as GP trees are evolved in this section.

\subsection{Evolutionary Model and the Fitness Assignment Process}\label{real_fit}

Our aim is to find a test function for which a given algorithm $A$ performs 
better than another given algorithm $B$. The test function that is being 
searched for will be evolved by using Genetic Programming \cite{koza1} with steady 
state \cite{syswerda1}.

The quality of the test function encoded in a GP chromosome is computed in a 
standard manner. The given algorithms $A$ and $B$ are applied to the test 
function. These algorithms will try to optimize (find the minimal value of) 
that test function. To avoid the lucky guesses of the optimal point, each 
algorithm is run 500 times and the results are averaged. Then, the fitness 
of a GP chromosome is computed as the difference between the averaged 
results of the algorithm $A$ and the averaged results of the algorithm $B$. In the 
case of function minimization, a negative fitness of a GP chromosome means 
that the algorithm $A$ performs better than the algorithm $B$ (the values obtained 
by $A$ are smaller (on average) than those obtained by $B)$.

\subsection{Algorithms Used for Comparison}\label{real_acceptance}

We describe several evolutionary algorithms used for comparison purposes. 
All the algorithms described in this section are embedded in the NFL-style 
algorithm described in section \ref{nfl}. More precisely, the considered algorithms 
particularize the solution representation, the mutation operator, and the 
acceptance mechanism (the procedure \textit{Accepted}) of the NFL algorithm described in 
section \ref{nfl}. The mutation operator is the only search operator used for 
exploring the neighborhood of a point in the search space.\\

\textbf{\textit{A}}$_{1}$ -- real encoding (the individuals are represented 
as real numbers using 32 bits), Gaussian mutation with $\sigma _{1}$ = 
0.001, the parent and the offspring compete for survival.\\

\textbf{\textit{A}}$_{2}$ -- real encoding (the individuals are represented 
as real numbers using 32 bits), Gaussian mutation with $\sigma _{2}$ = 
0.01, the parent and the offspring compete for survival.\\

\textbf{\textit{A}}$_{3}$ -- binary encoding (the individuals are 
represented as binary strings of 32 bits), point mutation with $p_{m}$ = 0.3, 
the parent and the offspring compete for survival.\\

\textbf{\textit{A}}$_{4}$ -- binary encoding (the individuals are 
represented as binary strings of 32 bits), point mutation with $p_{m}$ = 0.1, 
the parent and the offspring compete for survival.

\subsection{Numerical Experiments}\label{real_experiments}

Several numerical experiments for evolving functions matched to a given 
algorithm are performed in this section. The algorithms used for comparison 
have been described in section \ref{real_acceptance}.

The number of dimensions of the space is set to 1 (i.e. one-dimensional 
functions) and the definition domain of the evolved test functions is [0, 1].

The parameters of the GP algorithm are given in Table \ref{real_tab2}.

\begin{table}[htbp]
\caption{The parameters of the GP algorithm used for numerical experiments}
\label{real_tab2}
\begin{center}
\begin{tabular}
{p{140pt}p{135pt}}
\hline
\textbf{Parameter}& 
\textbf{Value} \\
\hline
Population size& 
50 \\
Number of generations& 
10 \\
Maximal GP tree depth& 
6 \\
Function set& 
$F$ = {\{}+, -, *, \textit{sin}, \textit{exp}{\}} \\
Terminal set& 
$T$ = {\{}$x${\}} \\
Crossover probability& 
0.9 \\
Mutation& 
1 mutation / chromosome \\
Runs& 
30 \\
\hline
\end{tabular}
\end{center}
\end{table}

The small number of generations (only 10) has been proved to be sufficient for 
the experiments performed in this paper.

Evolved functions are given in Table \ref{real_evolved}. For each pair ($A_{k}$, $A_{j})$ is 
given the evolved test function for which the algorithm $A_{k}$ performs 
better than the algorithm $A_{j}$. The mean of the fitness of the best GP 
individual over 30 runs is also reported.

\begin{table}[htbp]
\caption{The evolved test functions}
\label{real_evolved}
\begin{center}
\begin{tabular}
{p{64pt}p{124pt}p{98pt}}
\hline
\textbf{Algorithms}& 
\textbf{Evolved Test Function}& 
\textbf{Averaged fitness} \\
\hline
($A_{1}$, $A_{2})$& 
$f_1 (x) = 0.$& 
0 \\
($A_{2}$, $A_{1})$& 
$f_2 (x) = - 6x^3 - x.$& 
-806.03 \\
($A_{3}$, $A_{4})$& 
$f_3 (x) = x - 2x^5.$& 
-58.22 \\
($A_{4}$, $A_{3})$& 
$f_4 (x) = - 4x^8.$& 
-34.09 \\
($A_{2}$, $A_{4})$& 
$f_5 (x) = 0.$& 
0 \\
($A_{4}$, $A_{2})$& 
$f_6 (x) = - 6x^3 - x.$& 
-1601.36 \\
\hline
\end{tabular}
\end{center}
\end{table}

From Table \ref{real_evolved} it can be seen that the proposed approach made possible the 
evolving of test functions matched to the most of the given algorithms. The 
results of these experiments give a first impression of how difficult the 
problems are. Several interesting observations can be made:

The GP algorithm was able to evolve a function for which the algorithm 
$A_{2}$ (real encoding with $\sigma $ = 0.01) was better then the algorithm 
$A_{1}$ (real encoding with $\sigma $ = 0.001) in all the runs (30). However, 
the GP algorithm was not able to evolve a test function for which the 
algorithm $A_{1}$ is better that the algorithm $A_{2}$. In this case the 
function $f(x)$ = 0 (where both algorithms perform the same) was the only one to 
be found. It seems to be easier to find a function for which an algorithm 
with larger "jumps" is better than an algorithm with smaller "jumps" 
than to find a function for which an algorithm with smaller "jumps" is 
better than an algorithm with larger "jumps".

A test function for which the algorithm $A_{4}$ (binary encoding) is better 
than the algorithm $A_{2}$ (real encoding) was easy to find. The reverse 
(i.e. a test function for which the real encoding algorithm $A_{2}$ is better 
than the binary encoded algorithm $A_{4})$ has not been found by using the GP 
parameters considered in Table \ref{real_tab2}.

\section{Binary-valued functions}\label{binary_representation}

Test functions represented as binary strings are evolved in this section. We employed the binary-strings representation for the test functions because in this way we can evolve any function without being restricted to a given set of operators.

\subsection{Prerequisite}\label{pre}

Our analysis is performed in the finite search space $X$ \cite{wolpert1}. The space of possible "cost" values, $Y$, is also finite. We restrict our analysis to binary search spaces. This is not a hard restriction since all other values can be represented as binary strings. Thus $X = \{0,1\}^{n}$ and $X = \{0,1\}^{m}$. 

An optimization problem $f$ is represented as a mapping $f:X \mapsto Y$. 

The set $F=Y^X$ denotes the space of all possible problems. The size of $F$ is $|Y|^{|X|}$.

In our experiments $n=16$ and $m=8$. Thus $|X|=2^{16}=65536$ and $|Y|=2^{8}=256$. The number of optimization problems in this class is $|Y|^{|X|}=256^{65536} \approx 10^{157826}$. 

Each test problem in this class can be stored in a string of $65536*8 bits = 512Kb$.

Within this huge search space we will try to find test problems for which a given algorithm $A$ is better than another given algorithm $B$.

\subsection{Evolutionary Model and the Fitness Assignment Process}\label{binary_fit}

Our aim is to find a test function for which a given algorithm $A$ performs 
better than another given algorithm $B$. The test function that is being 
searched for will be represented as strings over the \{0,1\} alphabet as described in section \ref{pre}. 

The algorithm used for evolving these functions is a standard steady state \cite{syswerda1} Evolutionary Algorithm that works with a binary encoding of individuals \cite{back1,back2}. Each test problem in this class can be stored in a string of $65536*8 bits = 512Kb$. 

The most important aspect of this algorithm regards the way in which the fitness of an individual is computed. 

The quality of the test function encoded in a chromosome is computed as follows: The given algorithms $A$ and $B$ are applied to the test function. These algorithms will try to optimize (find the minimal value of) 
that test function. To avoid the lucky guesses of the optimal point, each 
algorithm is run 100 times and the results are averaged. Then, the fitness 
of a chromosome encoding a test function is computed as the difference between the averaged 
results of the algorithm $A$ and the averaged results of the algorithm $B$. 

In the case of function minimization, a negative fitness of a chromosome means 
that the algorithm $A$ performs better than the algorithm $B$ (the values obtained 
by $A$ are smaller (on average) than those obtained by $B)$.

\subsection{Algorithms Used for Comparison}\label{binary_acceptance}

We describe several evolutionary algorithms used for comparison purposes. 
All the algorithms described in this section are embedded in the NFL 
algorithm described in section \ref{nfl}. More precisely, the considered algorithms 
particularize the solution representation, the mutation operator, and the 
acceptance mechanism (the procedure \textit{Accepted}) of the NFL algorithm described in 
section \ref{nfl}. 

All algorithms are derived from (1+1) ES and can be described as follows:
\begin{itemize}

\item[{\it (i)}] Individuals are represented as binary strings over the search space \textit{X}.
\item[{\it (ii)}] Mutation operator is the only search operator used for 
exploring the neighborhood of a point in the search space.
\item[{\it (iii)}] The parent and the offspring compete for survival.
\end{itemize}

The number of mutations / chromosome is a parameter of the compared algorithms. The range for this parameter is 1 up to chromosome length. If the number of mutations / chromosome is equal to the chromosome length, the considered algorithm will behave like Random Search. 

Since the number of mutations / chromosome is different from algorithm to algorithm we denote by $B_{k}$ the NFL-style Algorithm that performs $k$ mutations / chromosome.

\subsection{Numerical Experiments}\label{binary_experiments}

Several numerical experiments for evolving functions matched to a given 
algorithm are performed in this section. The algorithms used for comparison 
have been described in section \ref{binary_acceptance}.

The parameters of the algorithm used for evolving test functions are given in Table \ref{binary_tab2}.

\begin{table}[htbp]
\caption{The parameters of the algorithm used for evolving test functions}
\label{binary_tab2}
\begin{center}
\begin{tabular}
{p{110pt}p{110pt}}
\hline
\textbf{Parameter}& 
\textbf{Value} \\
\hline
Population size& 
10 \\
Number of generations& 
10 \\
Crossover type&
Uniform\\
Crossover probability& 
0.9 \\
Mutation type&
Point mutation\\
Mutation probability& 
0.01 \\
Chromosome length& 
65536*8 bits \\
Runs& 
30 \\
\hline
\end{tabular}
\end{center}
\end{table}

The small number of generations (only 10) has been proved to be sufficient for 
the experiments performed in this paper.

Results are given in Table \ref{binary_tab3}. For each pair ($B_{k}$, $B_{j})$ is 
given the average (over 30 runs) of best fitness scored by an individual (encoding a test function) for which the algorithm $B_{k}$ performs better than the algorithm $B_{j}$.

\begin{table}[htbp]
\caption{Fitness of the best individual in the last generation. Results are averaged over 30 independent runs.}
\label{binary_tab3}
\begin{center}
\begin{tabular}
{|p{17pt}|p{17pt}|p{17pt}|p{17pt}|p{17pt}|p{17pt}|p{17pt}|p{17pt}|p{17pt}|p{17pt}|p{17pt}|p{17pt}|p{17pt}|p{17pt}|p{17pt}|p{17pt}|p{17pt}|}
\hline
& 
$B_{1}$& 
$B_{2}$& 
$B_{3}$& 
$B_{4}$& 
$B_{5}$& 
$B_{6}$& 
$B_{7}$& 
$B_{8}$& 
$B_{9}$& 
$B_{10}$& 
$B_{11}$& 
$B_{12}$& 
$B_{13}$& 
$B_{14}$& 
$B_{15}$& 
$B_{16}$ \\
\hline
$B_{1}$& 
-& 
-95 & 
-53 & 
-19 & 
-102 & 
-69 & 
-79 & 
-108 & 
-151 & 
-214 & 
-182 & 
-192 & 
-159 & 
-246 & 
-259 & 
-284  \\
\hline
$B_{2}$& 
-331& 
-& 
-132 & 
-119 & 
-195 & 
-177 & 
-197 & 
-163 & 
-192& 
-230& 
-254 & 
-284& 
-261& 
-291& 
-313& 
-356 \\
\hline
$B_{3}$& 
-454& 
-283 & 
-& 
-122 & 
-196& 
-179& 
-171& 
-234& 
-264& 
-293& 
-236& 
-319& 
-317& 
-324& 
-315& 
-333  \\
\hline
$B_{ 4}$& 
-495& 
-332& 
-168& 
-& 
-176& 
-159& 
-190& 
-185& 
-248& 
-259& 
-327& 
-321& 
-330& 
-421& 
-334& 
-374  \\
\hline
$B_{ 5}$& 
-496& 
-309& 
-164& 
-132& 
-& 
-125& 
-198& 
-151& 
-266& 
-244& 
-285& 
-250& 
-259& 
-346 & 
-364& 
-338  \\
\hline
$B_{ 6}$& 
-555& 
-286& 
-169& 
-126& 
-180& 
-& 
-175& 
-163& 
-200& 
-243& 
-197& 
-284& 
-276& 
-340& 
-364& 
-351  \\
\hline
$B_{ 7}$& 
-531& 
-358& 
-198& 
-140& 
-149& 
-138& 
-& 
-159& 
-199& 
-250& 
-207& 
-230& 
-262& 
-272& 
-306& 
-345  \\
\hline
$B_{ 8}$& 
-513& 
-323& 
-196& 
-159& 
-160& 
-140& 
-174& 
-& 
-202& 
-220& 
-217& 
-259& 
-289& 
-311& 
-291 & 
-328  \\
\hline
$B_{ 9}$& 
-538& 
-338 & 
-201& 
-156& 
-138& 
-145& 
-153& 
-150& 
-& 
-204& 
-197& 
-237& 
-271& 
-306& 
-312 & 
-315 \\
\hline
$B_{ 10}$& 
-526& 
-336& 
-190& 
-190& 
-156& 
-134& 
-149& 
-151& 
-131& 
-& 
-188& 
-187& 
-245& 
-228& 
-274& 
-292 \\
\hline
$B_{ 11}$& 
-528& 
-321& 
-210& 
-161 & 
-169& 
-136 & 
-147& 
-131& 
-156& 
-190& 
-& 
-245& 
-248& 
-275& 
-294& 
-325 \\
\hline
$B_{ 12}$& 
-602& 
-296& 
-224& 
-191& 
-145& 
-148& 
-121& 
-148 & 
-122& 
-166& 
-115& 
-& 
-205& 
-234& 
-262& 
-287  \\
\hline
$B_{ 13}$& 
-493& 
-331& 
-193& 
-127 & 
-152& 
-136& 
-125& 
-167& 
-143& 
-182 & 
-154& 
-164& 
-& 
-226& 
-238 & 
-236 \\
\hline
$B_{ 14}$& 
-544& 
-345& 
-210& 
-181& 
-174& 
-149 & 
-117& 
-137& 
-125& 
-158& 
-123& 
-147& 
-158& 
-& 
-175& 
-218 \\
\hline
$B_{ 15}$& 
-543& 
-298& 
-212 & 
-150& 
-128& 
-103& 
-109& 
-89 & 
-126& 
-148& 
-153& 
-181& 
-150& 
-195& 
-& 
-202  \\
\hline
$B_{ 16}$& 
-583& 
-329& 
-234& 
-151& 
-144& 
-129& 
-108& 
-83 & 
-115& 
-134& 
-147& 
-158& 
-129& 
-161& 
-173& 
- \\
\hline
\end{tabular}
\end{center}
\end{table}

Table \ref{binary_tab3} shows that the proposed approach made possible the 
evolving of test functions matched to all given algorithms (all fitness values are negative). The 
results of these experiments give a first impression of how difficult the 
problems are. Several interesting observations can be made:

In the first row of data (corresponding to the algorithm $B_{1}$) the average fitness decrease from -95 (for the pair ($B_{1}, B_{2}$)) to -284 (for the pair ($B_{1}, B_{16}$)). Knowing that a negative value of the fitness means that the algorithm $B_{1}$ is better than the algorithm $B_{k}$ we may infer that is more easy to find a function for which an algorithm performing 1 mutation/chromosome is better than an algorithm performing 16 mutations/chromosome (the algorithm $B_{16}$ which is actually behaves like Random Search) than to find a test function for which an algorithm performing 1 mutation/chromosome is better than an algorithm performing 2 mutations/chromosome.

If we take a look at each row of data after the cells in the first diagonal we can see that the values have an descending tendency. This means that is easier to beat an algorithm performing more mutations than to beat an algorithm performing less mutations / chromosome.

In the first column the values have a descending trend, too. This means that in the space of test functions it is more easy to find a test function for which an algorithm performing 16 mutations/chromosome is better an algorithm performing 1 mutation/chromosome than to find a test function for which an algorithm performing 2 mutations/chromosome is better an algorithm performing 1 mutation/chromosome. This results can be explained by the fact that $B_{1}$ and $B_{16}$ are very different whereas $B_{1}$ and $B_{2}$ are very similar is more  difficult to find test problems for which two very similar algorithms perform significantly  different.

The lowest value in Table \ref{binary_tab3} is -583 and it corresponds to pair ($B_{16}, B_{1}$). This means that finding a function for which an algorithm performing 16 mutations/chromosome is better than an algorithm performing 1 mutation/chromosome was the easiest operation. This suggests a rugged fitness landscape of the evolved test function. In order to confirm this hypothesis we have analyzed the landscape of the evolved test functions (30 functions obtained in 30 runs) for the pair ($B_{16}, B_{1}$). Each test function was considered as having 1 real-valued variable over the interval [0, 65535]. The average number of peaks (a point is considered as being a peak (local or global optimum) if its left and right values are higher than its value) of the evolved test functions was 23127 (out of 65536 points). This suggests an highly rugged landscape.

\section{Conclusions and Further Work}

In this paper, a framework for evolving test functions that are matched to a 
given algorithm has been proposed. The proposed framework is intended to 
provide a practical evidence for the NFL theories. Numerical experiments 
have shown the efficacy of the proposed approach: test functions for which Random Search performs better than all other considered evolutionary algorithm have been successfully evolved.

Further research will be focused on the following directions:

\begin{itemize}

\item[{\it (i)}] Proving that the evolved test problems are indeed hard for the considered Evolutionary Algorithms.

\item[{\it (ii)}] Comparing other evolutionary algorithms for single and multiobjective 
optimization. Several test functions matched to some classical algorithms 
(such as standard GAs or ES) for function optimization will be evolved. In 
this case the problem is much more difficult since the number of distinct 
solutions visited during the search process could be different for each 
algorithm.

\item[{\it (iii)}] Evolving difficult test instances for algorithms used for solving other real-world problems (such as TSP \cite{back2}, symbolic regression \cite{koza1}, classification etc).

\item[{\it (iv)}] Finding the set (class) of test problems for which an algorithm is better than the other.

\end{itemize}


\begin{thebibliography}{50}

\bibitem{back1}
Back, T.: Evolutionary Algorithms in Theory and Practice. Oxford University Press, Oxford (1996)
%
\bibitem{back2}
Back, T., Fogel, D.B., Michalewicz, Z. (eds.): Handbook of Evolutionary Computation. Institute of Physics Publishing, Bristol, and Oxford University Press, New York (1997)
%
\bibitem{deb1}
Deb, K.: Multi-Objective Genetic Algorithms: Problem Difficulties and Construction of Test Functions, Evolutionary Computation, MIT Press, Cambridge, MA, Vol. \textbf{7}, (1999) 205-230
%
\bibitem{droste1}
Droste, S., Jansen, T.,  Wegener, I.: A Natural and Simple Function which is Hard for All Evolutionary Algorithms, technical report, 2000, Univ. Dortmund, Germany, (2000)
%
\bibitem{droste2}
Droste, S., Jansen, T., Wegener, I.: Perhaps not a Free Lunch but at Least a Free Appetizer, In: Banzhaf W. (Eds.), Proceedings of the Genetic and Evolutionary Computation Conference,  Morgan Kaufmann (1999) 833-839
%
\bibitem{igel1}
Igel, C. Toussaint, M.: On Classes of Functions for which No Free Lunch
Results Hold, see http://citeseer.nj.nec.com/528857.html, (2001)
%
\bibitem{glover1}
Glover, F.: Tabu search - Part I, ORSA Journal of Computing, \textbf{1} (1989) 190-206
%
\bibitem{glover2}
Glover, F.: Tabu search - Part II, ORSA Journal of Computing, \textbf{2} (1990) 4-32
%
\bibitem{goldberg1}
Goldberg, D.E.: Genetic Algorithms in Search, Optimization, and Machine Learning, Addison-Wesley, Reading, MA (1989)
%
\bibitem{koza1}
Koza, J. R.: Genetic Programming: On the Programming of Computers by Means of Natural Selection, MIT Press, Cambridge, MA, (1992)
%
\bibitem{knowles1}
Knowles, J. D. Corne, D.W.: Approximating the Nondominated Front using the Pareto Archived Evolution Strategies, Evolutionary Computation, MIT Press, Cambridge, MA, \textbf{8} (2000) 149-172
%
\bibitem{streeter1}
Streeter, M. J.: Two Broad Classes of functions for Which a No Free Lunch Result does not Hold, In Proceedings of the GECCO 2003, edited by Erick Cantu-Paz (et al), Springer-Verlag, Berlin (2003) 1418-1430
%
\bibitem{syswerda1}
Syswerda, G.: Uniform Crossover in Genetic Algorithms, In: Proceedings of the 3$^{rd}$ International Conference on Genetic Algorithms, Schaffer J.D. (Editor), Morgan Kaufmann Publishers, San Mateo, CA (1989) 2-9
%
\bibitem{yao1}
Yao, X., Liu, Y., Lin, G.: Evolutionary Programming Made Faster, IEEE Transaction on Evolutionary Computation, \textbf{3}, (1999) 82-102
%
\bibitem{whitley1}
Whitley, L.D.: A Free Lunch Proof for Gray versus Binary Encodings, In Banzhaf, W., (et. al) (eds.) Proceedings of the Genetic and Evolutionary Computation Conference, Vol. 1, Morgan Kaufmann, (1999) 726-733
%
\bibitem{whitley2}
Whitley, L.D.,: Functions as Permutations: Implications for No Free Lunch, Walsh
Analysis and Statistics, In: Schoenauer, M., (et. al) (eds.). Proceedings of the Sixth International Conference on
Parallel Problem Solving from Nature (PPSN VI), Springer-Verlag, Berlin, (2000) 169-178
%
\bibitem{wolpert1}
Wolpert, D.H., McReady, W.G.: No Free Lunch Theorems for Optimization, IEEE Transaction on Evolutionary Computation, \textbf{1} (1997) 67-82
%
\bibitem{wolpert2}
Wolpert, D.H., McReady, W.G.: No Free Lunch Theorems for Search, Technical Report, SFI-TR-05-010, Santa Fe Institute, (1995)

\end{thebibliography}
\end{document}